\documentclass[10pt,twocolumn,letterpaper]{article}
\pdfoutput=1

\usepackage{iccv}
\usepackage{times}
\usepackage{epsfig}
\usepackage{graphicx}
\usepackage{amsmath}
\usepackage{amssymb}

\usepackage{todonotes}
\usepackage{booktabs}
\usepackage[font=small]{caption}
\usepackage[font=small]{subcaption}
\usepackage{pgfplots}
\pgfplotsset{compat=1.9}
\usepgfplotslibrary{external}
\usepgfplotslibrary{fillbetween}
\usepackage{algorithm}
\usepackage[noend]{algpseudocode}
\usepackage{lipsum}

\usepackage[pagebackref=true,breaklinks=true,letterpaper=true,colorlinks,bookmarks=false]{hyperref}

\usepackage[capitalize]{cleveref}
\crefname{section}{Sec.}{Secs.}
\Crefname{section}{Section}{Sections}
\Crefname{table}{Table}{Tables}
\crefname{table}{Tab.}{Tabs.}
\crefname{alg}{alg.}{algs.}
\Crefname{alg}{Algorithm}{Algorithms}

\iccvfinalcopy 

\def\iccvPaperID{11} 

\ificcvfinal\fi

\begin{document}

\title{Benchmarking Data Efficiency and Computational Efficiency \\ of  Temporal Action Localization Models}

\newcommand{\jvg}[1]{[{\textbf{ \color{blue} JvG: #1}}]}


\author{
Jan Warchocki$^{*}$ \quad Teodor Oprescu$^{*}$ \quad Yunhan Wang\thanks{Equal contribution} \quad Alexandru D\u{a}m\u{a}cu\c{s} \quad Paul Misterka \\ Robert-Jan Bruintjes \quad Attila Lengyel \quad Ombretta Strafforello \quad Jan van Gemert\\
Computer Vision Lab, Delft University of Technology\\
Delft, the Netherlands\\
}


\maketitle
\ificcvfinal\thispagestyle{empty}\fi

\begin{abstract}
   In temporal action localization, given an input video, the goal is to predict which actions it contains, where they begin, and where they end. Training and testing current state-of-the-art deep learning models requires access to large amounts of data and computational power. However, gathering such data is challenging and computational resources might be limited. This work explores and measures how current deep temporal action localization models perform in settings constrained by the amount of data or computational power. We measure data efficiency by training each model on a subset of the training set. We find that TemporalMaxer outperforms other models in data-limited settings. Furthermore, we recommend TriDet when training time is limited. To test the efficiency of the models during inference, we pass videos of different lengths through each model. We find that TemporalMaxer requires the least computational resources, likely due to its simple architecture.
\end{abstract}


\section{Introduction}
\label{sec:introduction}

Temporal action localization (TAL) is concerned with automatically recognizing an action and its start and end in a video~\cite{survey_xia}. TAL has found potential use in domains such as video summarization \cite{video_summarization} and public video surveillance \cite{video_surveillance, survey_xia}. Various algorithms are proposed for TAL, and deep learning models such as such as TriDet \cite{tridet}, TemporalMaxer \cite{temporal_maxer}, and ActionFormer \cite{action_former} outperform models based on hand-crafted features \cite{survey_xia}. These deep learning models require large datasets to train on, such as THUMOS'14 \cite{thumos_14} or ActivityNet \cite{activity_net}. However, curating, annotating and storing  datasets of such scale is difficult, expenisve, and time-consuming \cite{survey_xia, fsl_tal, fsl_qat}. To save data, in this work we explore  data efficiency of deep learning-based TAL models.

In addition to data efficiency, we also evaluate compute efficiency. Compute efficiency is particularly relevant  when the success of Transformers \cite{transformers} in natural language processing (NLP) \cite{transformers, bert}, is employed in TAL~\cite{tad_tr, action_former}. Transformers are known to be computationally expensive \cite{efficient_transformers_survey, reformer}. To save computing resources, in this work, we explore how computationally efficient deep learning-based TAL methods are. 

Our analysis of data- and compute-efficiency focuses on ActionFormer \cite{action_former}, STALE \cite{stale}, TemporalMaxer \cite{temporal_maxer}, and TriDet \cite{tridet}, as they represent the current state-of-the-art in temporal action localization. The contributions of this paper are four-fold, as detailed below. 

First, we test the data efficiency of the TAL models. Inspired by Ding \etal \cite{kfc} and Henaff \cite{image_recognition_cpc}, we train each model multiple times on a percentage of the training set and report the average mean average precision (mAP). By applying this method on both THUMOS'14 \cite{thumos_14} and ActivityNet \cite{activity_net} datasets, we find that the TemporalMaxer \cite{temporal_maxer} performs the best in a data-limited setting. 

Second, we evaluate the effect of score fusion~\cite{untrimmednet, action_former, score_fusion_strat} on data efficiency. Score fusion combines the outputs of an evaluated model with the outputs of an auxiliary  model, often UntrimmetNet \cite{untrimmednet, action_former}. 
We find that score fusion can significantly increase the performances of the models. We thus recommend that when choosing a model for a custom dataset, the options both with and without score fusion should be considered.

Third, we test the computational efficiency of each model during training. We measure training performance by analyzing the trade-off between training time and obtained average mAP. We find that the TriDet model \cite{tridet} is the best choice in training time-limited settings, because it requires the least amount of training time but still obtains the best average mAP.

Fourth, we test the computational efficiency of each model during inference. We expand on the approach of measuring the computational complexity of the model by passing to it a video of a specific size \cite{action_former, tridet, temporal_maxer}. We evaluate each model on videos of increasing lengths and report the number of floating point operations, the memory consumed, and the inference time. We find that TemporalMaxer requires the least computational resources, while STALE~\cite{stale} requires the most. 

\section{Related work}
\label{sec:related_work}

\textbf{Action recognition.} The survey by Xia and Zhan \cite{survey_xia} identifies five different tasks in video understanding: untrimmed video classification, trimmed action recognition, temporal action proposals, temporal action localization, and dense-captioning events in videos. This work focuses on temporal action localization (TAL) for its potential uses in video summarization \cite{video_summarization} and public surveillance \cite{video_surveillance}. In TAL, the goal is to predict which actions happen in a video stream, where they begin, and where they end. The deep learning models created for this problem can be divided into two categories \cite{survey_xia}: two-stage and one-stage. Two-stage models \cite{tsa_net, bsn, bmn} attempt to first locate the actions and then classify them. One-stage models \cite{action_former, tridet, temporal_maxer, tad_tr} locate and classify the actions at the same time. This work analyzes ActionFormer \cite{action_former}, STALE \cite{stale}, TemporalMaxer \cite{temporal_maxer}, and TriDet \cite{tridet}, all of which are one-stage models.


\textbf{Testing for data efficiency.} This problem involves assessing how well a given model performs with limited training data available. A common approach is to use $n$-shot learning \cite{fsl_qat, fsl_tal}, which involves training the model on only $n$ samples per class. However, since a single class can be represented multiple times in a single video \cite{thumos_14, activity_net}, it is unclear whether $n$ should refer to the number of videos the given class appears in or whether it is the total number of instances of the class. Furthermore, representing each class equally would be difficult as the number of instances of a class per video varies. An alternative approach involves training on a given percentage $p$ of the samples from the training dataset \cite{image_recognition_cpc, kfc}. In this work, we use this approach.

\textbf{Optimizing for data efficiency.} As collecting and annotating datasets is expensive \cite{survey_xia}, related works have proposed few-shot TAL methods \cite{fsl_tal, fsl_qat}. These models use meta-learning and require all of the support videos to be input into the model at once. This makes their architecture incompatible with the architecture of current state-of-the-art models, which only expect a single video as input \cite{action_former, tridet, temporal_maxer}. This work, therefore, analyzes the data efficiency of some of the current state-of-the-art models.

\textbf{Testing for computational efficiency.} The term `computational efficiency` is often used to mean the number of floating point operations \cite{efficient_transformers_survey, transformer_in_transformer, action_former, tridet, temporal_maxer}, the memory used \cite{efficient_transformers_survey, reformer}, or the training \cite{compress_transformers} or inference time \cite{action_former, temporal_maxer, tridet}. In the task of temporal action localization, TriDet \cite{tridet}, TemporalMaxer \cite{temporal_maxer}, and ActionFormer \cite{action_former} all report the number of floating point operations as the amount of multiply-accumulate (MAC) operations and the time it takes to forward a single video of a fixed length through the model. However, no experiments have been performed that would show how these models scale with an increase in video length. This is relevant, as models that scale linearly, will asymptotically outperform models that scale \eg quadratically. Hence, even if a quadratic model outperforms a linear model on short videos, it will perform worse on longer videos. Thus, in this work, the inference performance of each of the tested models is measured on videos of increasing lengths.

Furthermore, motivated by \cite{compress_transformers}, this work reports the training time and the achieved mean average precision of each of the TAL models. This is done to better understand the suitability of each model for settings where the training time is limited.

\textbf{Optimizing for computational efficiency.} Both TriDet \cite{tridet} and TemporalMaxer \cite{temporal_maxer} aim to lower the required computational cost of ActionFormer \cite{action_former}. In TriDet, this is achieved by replacing the multi-head self-attention module with an efficient Scalable-Granularity Perception layer \cite{tridet}. TemporalMaxer, on the other hand, replaces the entire transformer module with a max-pooling block \cite{temporal_maxer}. This work compares the computational efficiencies of ActionFormer, STALE \cite{stale}, TemporalMaxer, and TriDet.

\section{Models}
\label{sec:models}

\textbf{ActionFormer.} ActionFormer~\cite{action_former} was one of the first models that showed a successful use of Transformers~\cite{transformers} in temporal action localization. The model uses an encoder-decoder architecture with a Transformer encoder and a convolutional decoder. At the time of its proposal, the model reached state-of-the-art performance on the THUMOS'14 dataset obtaining an average mAP of 66.8\%. The model also showed promising results on both the ActivityNet~\cite{activity_net} and EPIC-Kitchens 100~\cite{epic_kitchens} datasets. We also selected this model for evaluation, as the architectures of newer models, TriDet~\cite{tridet} and TemporalMaxer~\cite{temporal_maxer}, are inspired by the architecture of the ActionFormer.

\textbf{STALE.} \textit{Zero-\underline{S}hot \underline{T}emporal \underline{A}ction Detection via Vision-\underline{L}anguag\underline{e} Prompting} (STALE)~\cite{stale} is the most recent and state-of-the-art method in zero-shot temporal action localization. Inspired by CLIP~\cite{clip}, STALE uses a temporal vision transformer~\cite{temporal_vision_transformer} to encode videos into video embeddings and a text transformer~\cite{transformers} to encode class prompts into text embeddings. STALE attempts to learn an inter-relationship of vision-language via cross attention~\cite{transformers}. The model achieved average mAP of 52.9\% and 36.4\% on the THUMOS'14 and ActivityNet datasets respectively, outperforming similar models. We selected this model for evaluation, to compare it against methods that were not designed for a zero-shot learning scenario.

\textbf{TemporalMaxer.} The TemporalMaxer~\cite{temporal_maxer} model was constructed to require a low computational cost without sacrificing localization performance. Instead of employing a computationally-heavy backbone, such as a Transformer~\cite{transformers, action_former}, the model uses a basic, parameter-free max pooling block on top of a pre-trained 3D CNN. This model currently represents the state-of-the-art on the MultiTHUMOS dataset~\cite{multithumos} obtaining an average mAP of 29.9\%. Importantly, the model also has a lower computational complexity compared to other models. On a video of a length of around 5 minutes from the THUMOS'14 dataset, the inference time of the TemporalMaxer was observed to be 3x shorter than that of the ActionFormer. 

\textbf{TriDet.} The TriDet model~\cite{tridet} bases its architecture on the ActionFormer. Instead of using a multi-head self-attention mechanism, the model replaces it with an efficient Scalable-Granularity Perception (SGP) layer. The resulting model improves on the performance of the ActionFormer, obtaining an average mAP of 69.3\% on the THUMOS'14 dataset. Furthermore, the TriDet model represents the current state-of-the-art for the EPIC-Kitchens 100 dataset. Finally, the model was also shown to require less time and fewer floating point operations than the ActionFormer when performing inference on a 5 minute video from the THUMOS'14 dataset.

\section{Evaluation setup}
\label{sec:methodology}


\subsection{Data efficiency}


\textbf{Evaluation metrics.} Following common practice \cite{survey_xia, tridet, action_former, tsp, temporal_maxer}, the models were evaluated by reporting the achieved mean average precision (mAP) on different tIoU thresholds. Intersection over union (tIoU) is a 1-dimensional temporal Jaccard similarity metric and is thus computed as the ratio of the intersection of the predicted and actual duration of an action to their union. Given a tIoU threshold $\mu$ and a class $c$, correct predictions are those, whose tIoU $\geq \mu$ and the predicted class is the class $c$. Precision is then the ratio of the number of correct predictions to the total number of made predictions for the class $c$. As there can be multiple videos for each class $c$, average precision is the average of the precisions obtained in each of those videos. Finally, mean average precision is the average AP over all of the classes $c$. Thus, in general, given a fixed tIoU threshold $\mu$, the higher the mAP, the better the model performs.

\textbf{Testing procedure.} In this setup, it is assumed that a dataset $\mathcal{D}$ has a predefined split into a training set $\mathcal{D}_{\text{train}}$ and a testing set $\mathcal{D}_{\text{test}}$. Following works by Ding \etal \cite{kfc} and Henaff \cite{image_recognition_cpc}, a percentage $p$ of the training set $\mathcal{D}_{\text{train}}$ was randomly and uniformly sampled to create a subset $\mathcal{D}_\text{s}$. The models were then trained on the set $\mathcal{D}_\text{s}$ and evaluated on the set $\mathcal{D}_{\text{test}}$. During the evaluation, mean average precision  was calculated at different tIoU thresholds. The sampling, training, and testing procedure was repeated 5 times \cite{kfc, mixmatch} with different random splits. The mAP for each threshold was then averaged and the standard deviation was reported. The entire procedure was repeated for multiple percentages $p$. \Cref{alg:data_efficiency_testing} describes the exact testing procedure in the form of pseudocode.

In the pseudocode, the function $\operatorname{sample}$ randomly samples videos from the training set, such that:
\begin{equation}
    |\mathcal{D}_\text{s}| = \operatorname{round}\left(|\mathcal{D}_{\text{train}}| \cdot \frac{p}{100\%}\right)
\end{equation}
with $\operatorname{round}$ rounding the value to the nearest integer. Additionally, the function $\operatorname{sample}$ needs to ensure that each action class is represented at least once in the resulting set $\mathcal{D}_\text{s}$. In practice, this was realized by repeatedly sampling from the set $\mathcal{D}_{\text{train}}$ until a split, where all classes are represented, was found. The function $\operatorname{calculate-mAP}$ evaluates the model, that is, it calculates the mean average precision at different tIoU thresholds the model achieved on the test set $\mathcal{D}_{\text{test}}$. 

\begin{algorithm}[H]
    \begin{algorithmic}
        \State $\mathcal{D}_{\text{train}} = \{(\mathbf{X_i}, \mathbf{\hat{Y}_i})\}_{i=1}^N$
        \State $\mathcal{D}_{\text{test}} = \{(\mathbf{X_i}, \mathbf{\hat{Y}_i})\}_{i=1}^M$ 
        \For{$p = 10\%,\ldots,100\%$}
            \State mAPs $\gets$ empty list
            \For{$i = 1,\ldots,5$}
                \State $\mathcal{D}_{\text{s}} \gets \operatorname{sample}(\mathcal{D}_{\text{train}}, p)$
                \State $\text{Train on }\mathcal{D}_{\text{s}}$
                \State mAP $\gets \operatorname{calculate-mAP}(\mathcal{D}_{\text{test}})$
                \State mAPs.append(mAP)
            \EndFor
            \State $\text{Report }\operatorname{avg}(\text{mAPs})\text{ and }\operatorname{std}(\text{mAPs})$
        \EndFor
    \end{algorithmic}
    \caption{Data efficiency testing procedure}\label{alg:data_efficiency_testing}
\end{algorithm}
\vspace{-20pt}
\begin{figure}[H]
    \small{Algorithm 1. The data efficiency testing procedure. Assuming $\mathcal{D}_{\text{train}}$ and $\mathcal{D}_{\text{test}}$ are given, $\mathcal{D}_{\text{train}}$ is repeatedly subsampled with percentage $p$ to create the set $\mathcal{D}_\text{s}$. The model is then trained on $\mathcal{D}_\text{s}$ and evaluated on $\mathcal{D}_{\text{test}}$. The procedure is repeated 5 times for each percentage $p$, at each time reporting the averages of the mAPs and their standard deviation.}
\end{figure}

To understand the results between different datasets, for each percentage $p$ the expected number of instances per class is reported. This will help in investigating how many instances per class each model requires. Given a dataset $\mathcal{D}_{\text{train}}$ containing $N$ samples, having $M$ action instances in total, and $C$ action classes, the expected number of instances per class for each percentage $p$ is calculated as:
\begin{equation} \label{eq:instances_per_class}
    \text{\#/class} = \frac{p}{100\%} \cdot \frac{N}{C} \cdot \frac{M}{N} = \frac{p}{100\%} \cdot \frac{M}{C}
\end{equation}

It should be noted that the value computed in \Cref{eq:instances_per_class} is an estimate. The exact values would depend on the splits $\mathcal{D}_{\text{s}}$ used in the testing procedure. Nonetheless, this approximation was found to be useful in practice when comparing the models on different datasets.

\textbf{Score fusion.} Score fusion is a commonly used technique in TAL \cite{action_former, stale, tridet} to improve the performance of a model. Although the exact implementations vary between models, the general rule is that the final predictions made by a model are combined with the output of UntrimmedNet \cite{untrimmednet, action_former, stale, tridet}. UntrimmedNet \cite{untrimmednet} is a weakly-supervised action recognition model which only predicts video-level classes without temporal localization. It should be noted that UntrimmedNet is trained on the full ActivityNet and THUMOS datasets, respectively, while in practice limited training data would also apply to UntrimmedNet. Thus, in this work, the setups that use score fusion by default are also evaluated without score fusion.

\subsection{Computational efficiency}


\subsubsection{Training performance}

\hspace{\parindent}
Inspired by Li \etal \cite{compress_transformers}, the training time of each of the models is reported alongside the average mAP achieved on the test set. The training time is measured without initialization, that is, only the time spent in the training loop is measured. In this way, only the model performance is measured, without the time taken by external methods such as PyTorch data loaders. The training and testing procedure is repeated 5 times using different random seeds, each time measuring the time spent and the average mAP achieved.

\subsubsection{Inference performance}

\hspace{\parindent}

\textbf{Evaluation metrics.} Following the works on TriDet \cite{tridet}, TemporalMaxer \cite{temporal_maxer}, and ActionFormer \cite{action_former} the models were evaluated by reporting the total number of multiply-accumulate (MAC) floating point operations, memory consumed, and the inference time when fed an input video. To count the number of multiply-accumulate operations, the \texttt{fvcore} library \cite{fvcore} was used. As Transformer-based methods are known to require large amounts of memory \cite{efficient_transformers_survey, reformer}, we additionally report the total GPU memory (VRAM) footprint of each model, which is measured using the \texttt{max\_memory\_allocated} method from PyTorch. 

\textbf{Testing procedure.} The models were evaluated on randomly generated tensors, whose shapes correspond to videos of differing lengths. To guarantee the independence of results, the experiments for inference time, memory consumption, and number of MACs were run independently. Before each inference time measurement, the random tensor was passed through the model once as a warm-up procedure. Without this procedure, it was observed for the ActionFormer model that the inference time would be constant for all lengths of the input tensor. This was most likely caused by memory allocations happening as the tensor was being passed through the model. As such, the warm-up procedure was applied to all models to ensure a fair evaluation for all input sizes. Furthermore, the experiments for inference time were repeated 5 times \cite{tridet} with different random tensors. 

Additional setup was also required by the ActionFormer model. Given a dataset $\mathcal{D}$, the ActionFormer model is parameterized by a value \texttt{max\_seq\_len} indicating the maximum length of a video in $\mathcal{D}$ expressed in the number of features \cite{action_former}. During inference, all videos are padded with zeros to the \texttt{max\_seq\_len} length, which results in the same amount of computation done regardless of video length. To alleviate this issue, the value \texttt{max\_seq\_len} was configured to the lowest allowable value, which would be found through an inspection of the code. It should be noted that the value \texttt{max\_seq\_len} is only used during training and changing it during inference does not influence the output of the model, which was verified with one of the authors of the ActionFormer.

\section{Experiments}
\label{sec:experiments}

\begin{figure*}[ht]
    \begin{subfigure}[t]{0.45\linewidth}
        \centering
        \resizebox{\linewidth}{0.85\linewidth}{\centering \includegraphics[]{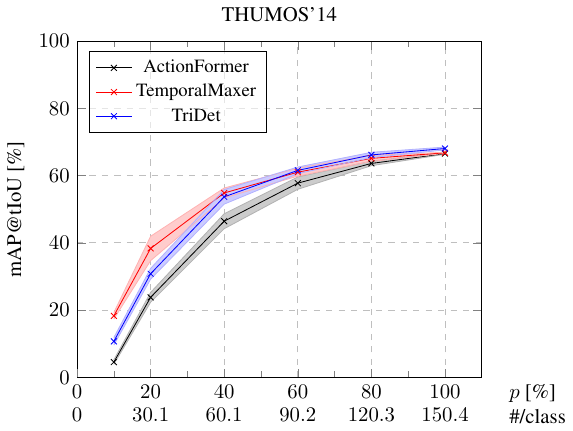}}
        \caption{Performance of the compared models on the THUMOS'14 dataset~\cite{thumos_14} in terms of average mAP@tIoU[0.3:0.1:0.7]. The TemporalMaxer model \cite{temporal_maxer} performs the best with little training data available, likely due to a simpler architecture. The TriDet model outperforms TemporalMaxer when the average number of instances per class is $> 100$.}
        \label{fig:results_data_efficiency_thumos}
    \end{subfigure}
    \hfill
    \begin{subfigure}[t]{0.45\linewidth}
        \resizebox{\linewidth}{0.85\linewidth}{\centering \includegraphics[]{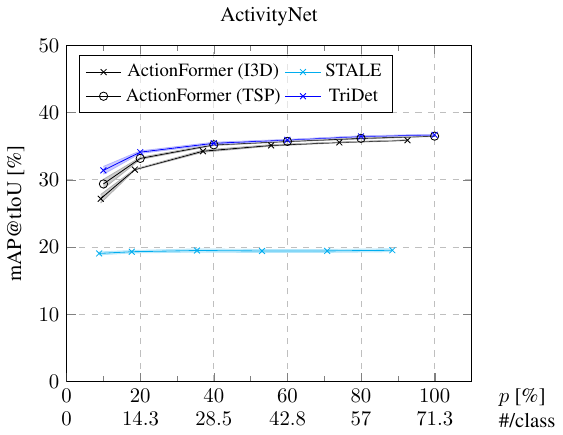}}
        \caption{Performance of the compared models on ActivityNet~\cite{activity_net} in terms of average mAP@tIoU[0.5:0.05:0.95]. The x-axis is shifted to the left for some setups due to fewer training samples being available. Most importantly, the ActionFormer and TriDet models can be seen to outperform the STALE model on all tested values of $p$. }
        \label{fig:results_data_efficiency_anet}
    \end{subfigure}
    \caption{Reported average mAP@tIoU for the tested models on the THUMOS'14 \cite{thumos_14} and ActivityNet \cite{activity_net} datasets. For each model, only the average mAP is shown. The width of each line corresponds to two standard deviations obtained by repeating the procedure 5 times for each $p$. Additionally, the expected average number of instances per class (\#/class) is reported as a secondary x-axis. We find that all of the models reach their near-best performance with less than or around 100 action instances per class.}
    \label{fig:results_data_efficiency}
\end{figure*}

\textbf{Datasets.} The models were evaluated on two datasets, commonly used to assess temporal action localization algorithms~\cite{action_former, tridet, stale}: THUMOS'14~\cite{thumos_14} and ActivityNet~\cite{activity_net}. THUMOS'14 contains 413 untrimmed videos and 20 action classes. This dataset is further split into a validation set, containing 213 videos and a test set containing 200 videos. In total, the validation set contains 3,007 action instances. We follow the configuration from the authors of the tested models and hence train the models on the validation set and test on the test set~\cite{action_former, tridet, temporal_maxer, stale}. ActivityNet contains around 20,000 videos with 200 action classes. The dataset is further split into a training set (10,024 videos), a validation set (4,926 videos), and a test set (5,044 videos). Using the approach from~\cite{action_former, stale, tridet}, the models are trained on the training set and evaluated on the validation set. As some of the videos from the ActivityNet dataset have become unavailable over time, it should be noted that the exact size of the dataset varies when using different models or features. 

\textbf{Features.} We take into consideration all features that were made available by the authors of a model for the given dataset. Hence, on the THUMOS'14 dataset, ActionFormer~\cite{action_former}, TemporalMaxer~\cite{temporal_maxer}, and TriDet~\cite{tridet} are all evaluated using the Inflated 3D (I3D) features~\cite{i3d_features}. On the ActivityNet dataset, the ActionFormer was evaluated using both I3D and TSP~\cite{tsp} features. STALE~\cite{stale} was tested with the CLIP~\cite{clip} features. The STALE model was not evaluated using I3D features due to limited compute availability. Finally, the performance of the TriDet model on the ActivityNet dataset was measured using the TSP features.

\textbf{Experimental setup.} All of the models were trained and tested using a single NVIDIA Tesla V100S 32GB located on an HPC cluster. All of the training and testing hyperparameters were left unchanged for the models unless otherwise stated in the subsequent sections. During data efficiency experiments, we therefore also use score fusion implemented by the ActionFormer, STALE, and TriDet models on the ActivityNet dataset~\cite{action_former, stale, tridet}. We reflect on the impact of the score fusion on the performance of the models in \Cref{sec:score_fusion}.

\subsection{Data efficiency}

\textbf{Results on THUMOS'14.} The results on the THUMOS'14 dataset can be seen in \Cref{fig:results_data_efficiency_thumos}. Firstly, we note that at $p = 100\%$, the average performance for each of the models is slightly lower than in the original works. We find an average mAP of $66.57 \pm 0.22~[\%]$ compared to $66.8\%$ for the ActionFormer, $66.79 \pm 0.16~[\%]$ contrary to $67.7\%$ for TemporalMaxer, and $68.07 \pm 0.42~[\%]$ instead of $69.3\%$ for TriDet. As noticed by~\cite{action_former}, however, different hardware setups may lead to different results, which might explain the differences observed in this work. Furthermore, we see that all models follow a similar learning curve. This is most likely caused by the fact that the models share a similar architecture, inspired by the architecture of the ActionFormer~\cite{action_former, tridet, temporal_maxer}. Moreover, as can be observed, at the low percentages $p$, the TemporalMaxer~\cite{temporal_maxer} model performs the best. This can be explained by the simpler architecture employed by the model, which would require less training data than the other models. We also note that for all models, the incline in performance noticeably slows down above $p = 60\%$, which corresponds to around $90$-$100$ action instances per class. We can thus see that each model saturates at around $100$ action instances per class and does not gain much from additional data. We also see that the TriDet model begins to outperform the TemporalMaxer around the same mark. 

\textbf{Results on ActivityNet.} As can be seen in \Cref{fig:results_data_efficiency_anet}, both the ActionFormer and the TriDet models outperform the STALE model on all tested percentages $p$. Furthermore, we observe that ActionFormer and TriDet saturate around the 40-60\% mark and do not gain from additional training data. This corresponds to around 30-40 action instances per class. We also notice that the STALE model does not visibly gain from an increase in training data. The model achieves an average mAP of $19.06 \pm 0.22~[\%]$ at $p = 10\%$ compared to $19.53 \pm 0.22~[\%]$ at $p = 100\%$. This flat learning curve is caused by the score enhancement, as is shown experimentally in \Cref{sec:score_fusion}. 

\textbf{Discussion.} From \Cref{fig:results_data_efficiency_thumos}, we can observe that the TemporalMaxer should likely be chosen in settings where the amount of data is limited. The simple architecture of that model allows it to show the best data efficiency out of the tested models. \Cref{fig:results_data_efficiency_anet} suggests that the ActionFormer or TriDet models should be chosen in favor of STALE. Based on the combined results in \Cref{fig:results_data_efficiency}, it is difficult to put an exact bound on the number of action instances per class required by the models. On both of the datasets, however, we can observe that the models reach their near-best performance with less than or around 100 action instances per class. This suggests making datasets larger will not further improve the performance of the tested models.

\subsubsection{Score fusion}
\label{sec:score_fusion}

\begin{figure}[h]
    \centering
    \resizebox{\linewidth}{0.85\linewidth}{\centering \includegraphics[]{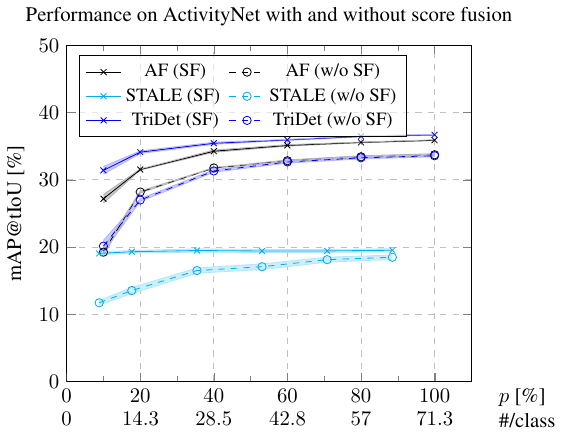}}
    \caption{Performance of ActionFormer (AF), STALE, and TriDet with score fusion (SF) and without score fusion (w/o SF). As we can observe, the performance of the models drops when score fusion is not used.}
    \label{fig:results_score_fusion}
\end{figure}

\hspace{\parindent}
In the default configuration, the score fusion techniques are used by the ActionFormer \cite{action_former}, STALE \cite{stale}, and TriDet \cite{tridet} on the ActivityNet dataset \cite{activity_net}. We repeat the data experiments without score fusion for these models on the ActivityNet dataset and report the results. We use the default features for these models for these experiments, hence, ActionFormer and TriDet use the TSP features \cite{tsp}, and STALE uses CLIP \cite{clip}. The results can be seen in \Cref{fig:results_score_fusion}. Score fusion improves the performance of the models for all tested values of $p$. The largest impact can be seen at low percentages $p$ in the small data regime.

\textbf{Discussion.} Unsurprisingly, score fusion has a significant influence on the performances of the models. It should be noted that score fusion is based on the assumption that UntrimmedNet class predictions are readily available, which in practice may not be the case. One should therefore be aware that performance on ActivityNet or Thumos does not always directly translate to true performance on a custom dataset, which may be lower. Alternatively, employing score fusion on custom data requires additional compute for retrieving the UntrimmedNet predictions. We argue that when choosing a model on a custom dataset, it is important to decide on the applicability of score fusion and evaluate the model both with and without score fusion.

\subsection{Computational efficiency}

\textbf{Concurrent jobs on the HPC cluster.} By default, the GPU nodes are shared between users in 
our compute cluster.
This setup could lead to a dependence of the training or inference time on the other jobs running on the cluster. To alleviate this issue, the training and inference time experiments were performed five times sequentially, such that the experiment jobs did not overlap. Therefore, the results for training and inference times are averaged over a total of 25 runs. We measure the remaining variance in training time, assuming that a low variance means that there are no important unmeasured confounding factors stemming from the concurrent use of the cluster. 

\subsubsection{Training efficiency}

\hspace{\parindent}
\textbf{Results.} We present the results in \Cref{tab:results_compute_training}. On THUMOS'14, TriDet achieves the best performance while requiring the least amount of training time on average. Interestingly, we find that the training time of the TemporalMaxer varies greatly between runs: from 1216.56 to 6829.95 seconds. This variance might come from the early stopping criterion employed in the training script of the model. Nonetheless, even in its fastest training run, TemporalMaxer is still the slowest of the tested models. On ActivityNet, ActionFormer and TriDet train for around five times as long as STALE, but also achieve much better performance. Finally, we note that the variance in training times was low for all models, except for the TemporalMaxer, thus the concurrent jobs on the cluster likely did not interfere with the experiment jobs.



\begin{table}[t]
    \centering
    \caption{Training performance of the compared models on the ActivityNet \cite{activity_net} and THUMOS'14 \cite{thumos_14} datasets. Both average training time and obtained average mAP are reported. On THUMOS'14, TriDet is the fastest and performs the best. On ActivityNet, the ActionFormer and TriDet models take longer to train than STALE but also achieve better performance.}  \label{tab:results_compute_training}
    \begin{tabular}{ccc}
        \toprule
        \textbf{Model} & \textbf{Time [s]} & \textbf{Avg. mAP [\%]}\\
        \midrule
        \multicolumn{3}{c}{\textbf{THUMOS'14}} \\
        \midrule
        AF & 887 $\pm$ 54 & 65.89 $\pm$ 0.09 \\
        TemporalMaxer & 2957 $\pm$ 1660 & 66.96 $\pm$ 0.37\\
        TriDet & 646 $\pm$ 26 & 68.07 $\pm$ 0.42\\
        \midrule
        \multicolumn{3}{c}{\textbf{ActivityNet}} \\
        \midrule
        ActionFormer (I3D) & 1945 $\pm$ 61 & 35.9 $\pm$ 0.14 \\
        ActionFormer (TSP) & 1932 $\pm$ 232 & 36.4 $\pm$ 0.05 \\
        STALE & 401 $\pm$ 6 & 19.37 $\pm$ 0.16 \\
        TriDet & 2236 $\pm$ 224 & 36.57 $\pm$ 0.18 \\
        \bottomrule
    \end{tabular}
\end{table}
\begin{figure*}[ht]
    \centering
    \begin{subfigure}[t]{0.33\linewidth}
        \centering
        \resizebox{\linewidth}{\linewidth}{\centering \includegraphics[]{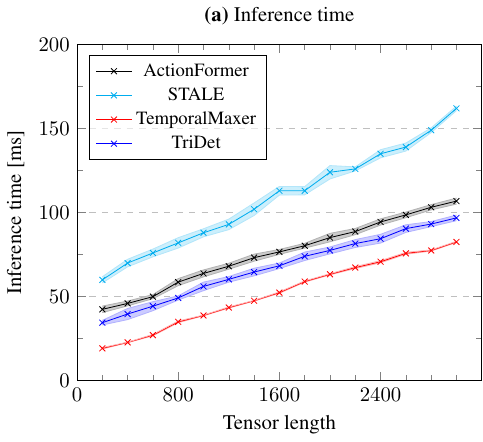}}
        \captionlistentry{}
        \label{fig:results_inference_time}
    \end{subfigure}
    \begin{subfigure}[t]{0.33\linewidth}
        \centering
        \resizebox{\linewidth}{\linewidth}{\centering \includegraphics[]{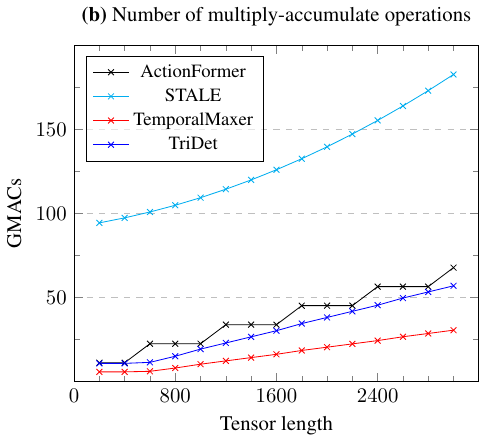}}
        \captionlistentry{}
        \label{fig:results_macs}
    \end{subfigure}
    \begin{subfigure}[t]{0.33\linewidth}
        \centering
        \resizebox{\linewidth}{\linewidth}{\centering \includegraphics[]{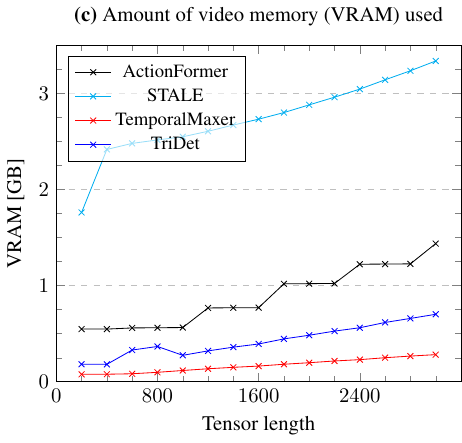}}
        \captionlistentry{}
        \label{fig:results_memory}
    \end{subfigure}
    \caption{Inference time, number of floating point operations, and memory consumption for ActionFormer 
    \cite{action_former}, TemporalMaxer \cite{action_former}, and STALE \cite{stale}. Most notably, we find that the TemporalMaxer model requires the least computational resources during inference, while STALE requires the most.}
    \label{fig:results_inference_efficiency}
\end{figure*}

\textbf{Discussion.} From the results obtained in \Cref{tab:results_compute_training} we find that the TriDet model should be chosen in settings where the training time is limited. This is due to the fact that the model was found to not only train for the least amount of time on THUMOS'14 but also achieve the best performance on both datasets. If the choice of the model is between ActionFormer and STALE, we find that the latter could be used in limited training time settings. Choosing STALE over the ActionFormer would, however, likely come with a decrease in TAL performance.


\subsubsection{Inference efficiency}

\hspace{\parindent}
\textbf{Additional experimental setup.} For the ActionFormer~\cite{action_former}, TemporalMaxer~\cite{temporal_maxer}, and TriDet~\cite{tridet} models, we obtained inference efficiency results by creating random tensors corresponding to I3D features~\cite{i3d_features} extracted from videos from the THUMOS'14 dataset~\cite{thumos_14}. We obtained results for the STALE model by creating random tensors corresponding to CLIP features~\cite{clip} on the ActivityNet dataset~\cite{activity_net}. The lengths of the tensors vary from 200 to 3000 in 200 increments. This range is dictated by the ActionFormer model, where the lowest allowable value of \texttt{max\_seq\_len} is 576, so videos of lengths longer than 3456 cannot be passed through the model without further changes to the configuration.

\textbf{Results.} As can be seen in \Cref{fig:results_inference_efficiency}, the TemporalMaxer model consistently achieves the lowest inference time, number of floating point operations, and memory consumption. This is because of the simple architecture of the model, which contains fewer parameters than other models \cite{temporal_maxer}. Conversely, we find that the STALE model is the most computationally expensive in all three tested metrics and on all tested lengths of the input video. Furthermore, we observe that the number of floating point operations and the memory consumption increase in steps for the ActionFormer model. This is because the model architecture requires padding input videos to multiples of 576. Nonetheless, the model scales linearly with respect to the input size. This thus matches the claims of the original work~\cite{action_former}. We see that TriDet and TemporalMaxer both also scale linearly with respect to the input size. As can be seen in \Cref{fig:results_macs}, the computational complexity of the STALE model does not increase linearly. A similar pattern is observable for memory consumption of STALE in \Cref{fig:results_memory}. Interestingly, we find a linear pattern in the inference time of STALE in \Cref{fig:results_inference_time}.

\textbf{Discussion.} In case of limited compute resources, TemporalMaxer should be chosen. TemporalMaxer requires the least amount of computational power on all tested video lengths. STALE should not be chosen in such settings, not only due to higher computational complexity but also because it scales non-linearly with respect to input video length. Hence, even if a configuration would be found that causes STALE to be more efficient on short videos, asymptotically it will still be worse than any other linear model.

\section{Conclusion}
\label{sec:conclusion}

In this work, we ask how well state-of-the-art temporal action localization models perform in settings limited by the amount of training data or computational resources available. We find that in a data deficient setting the TemporalMaxer model \cite{temporal_maxer} works the best, likely due to its simple architecture, which consists of fewer parameters compared to other models and does not use a Transformer backbone. Additionally, we find that performance barely improves when adding data beyond 100 action instances per class. This suggests making datasets larger will not further improve the performance of the tested models. The use of score fusion was shown to improve the performances of the models, hence when training a model on a custom dataset, options with and without score fusion should be considered. Furthermore, we test computational efficiency during training and inference. We find that TriDet \cite{tridet} offers the lowest training time as well as the best performance. Additionally, we find that TemporalMaxer requires the least computational resources at inference time, again likely due to its simple architecture without a Transformer backbone.

\textbf{Limitations.} It should be noted that the method for measuring data efficiency is limited as ActionFormer and TriDet are the only models that were evaluated on both datasets. Furthermore, the procedures for testing training and inference efficiency have limitations. The models have only been so far evaluated on the THUMOS'14 and ActivityNet datasets. The results on different datasets could lead to different conclusions. Furthermore, the timing experiments have been performed on a shared HPC cluster. It was however observed that the variance in training and inference times was small, which indicates that the concurrent jobs did not interfere with the experimental jobs. 

\textbf{Future work.} This work provides insights that will help in developing future data or computationally efficient TAL models. Based on the results of ActionFormer \cite{action_former} and STALE \cite{stale}, we see that self-attention should not be the mechanism of choice if the training data or computational resources are limited. We find that replacing such modules with custom layers, such as SGP \cite{tridet} or replacing transformer modules with max pooling \cite{temporal_maxer} improves the efficiency of the model. Finally, we note that future work in evaluating current models in terms of data or computational efficiency is possible. More models could be evaluated or the models could be evaluated on more datasets. 

\small
\smallskip\noindent\textbf{Acknowledgements.} 
This project is (partly) financed by the Dutch Research Council (NWO) (project VI.Vidi.192.100).

{\small
\bibliographystyle{ieee_fullname}
\bibliography{main}
}

\end{document}